\def\editmode{0}

\def\bibfilenames{bibman_refs}

\def\spsformat{0}

\if\spsformat1

\documentclass{article}
\usepackage{spconf}
\usepackage{comment}

\else

\documentclass[12pt,twocolumn]{IEEEtran}

\renewcommand{\baselinestretch}{.7184}

\fi

\usepackage{savesym} 
\savesymbol{labelindent}
\usepackage{enumitem} 
\newcommand{\acom}[1]{\textcolor{red}{{[#1]}}} 

\if\editmode1  
\usepackage[backend=bibtex,style=alphabetic,sorting=debug,maxbibnames=99]{biblatex}
\DeclareFieldFormat{labelalpha}{\thefield{entrykey}}
\DeclareFieldFormat{extraalpha}{}
\bibliography{\bibfilenames}
\newcommand{\cmt}[1]{\noindent\textcolor{lightgreen}{\underline{[#1]}}} 

\setlistdepth{9}
\newlist{bulletlist}{enumerate}{9}
\setlist[bulletlist,1]{label=$\bullet$}
\setlist[bulletlist,2]{label=$\diamond$}
\setlist[bulletlist,3]{label=$\rightarrow$}
\setlist[bulletlist,4]{label=$\circ$}
\setlist[bulletlist,5]{label=$-$}
\setlist[bulletlist,6]{label=$\square$}
\setlist[bulletlist,7]{label=$\star$}
\setlist[bulletlist,8]{label=$\checkmark$}
\setlist[bulletlist,9]{label=$\Delta$}
\newenvironment{bullets}{\begin{bulletlist}}{\end{bulletlist}}

\newcommand{\blt}[1][noargpassed]{
  \item%
  \ifthenelse{\equal{#1}{noargpassed}}{}{\cmt{#1}}%
}

\else
\usepackage{cite}
\newcommand{\cmt}[1]{} 
\newenvironment{bullets}{}{}
\newcommand{\blt}[1][noargpassed]{\ignorespaces}

\fi

\newcommand{\printmybibliography}{
\if\editmode1 
\printbibliography
\else
\bibliography{\bibfilenames}
\fi
}

\usepackage{fixltx2e}

\usepackage{graphicx}

\usepackage{subcaption}              

\usepackage[utf8]{inputenc}

\usepackage{amsfonts}
\usepackage{amsmath}
\usepackage{mathtools}

\usepackage{amssymb}

\usepackage{bm}

\usepackage{color,verbatim}
\usepackage{multirow}
\usepackage{accents}

\usepackage{theoremref}

\usepackage{url}

\newcounter{rulecounter}
\newcommand{\resetrule}{ \setcounter{rulecounter}{0}}
\resetrule

\newtheorem{myauxproblem}{Problem}

\newtheorem{myauxoptionalproblem}{Optional Problem}

\newsavebox{\selvestebox}
\newenvironment{colbox}[1]
  {\newcommand\colboxcolor{#1}%
   \begin{lrbox}{\selvestebox}%
   \begin{minipage}{\dimexpr\columnwidth-2\fboxsep\relax}}
  {\end{minipage}\end{lrbox}%
   \begin{center}
   \colorbox{\colboxcolor}{\usebox{\selvestebox}}
   \end{center}}

\definecolor{orange}{rgb}{1,0.8,0}
\definecolor{gray}{rgb}{.9,0.9,0.9}
\definecolor{darkgray}{rgb}{.3,0.3,0.3}
\definecolor{darkblue}{rgb}{.1,0.0,0.3}
\definecolor{lightblue}{rgb}{0.7,0.7,1}
\definecolor{lightred}{rgb}{1,0.7,.7}
\definecolor{purple}{RGB}{204,153,255}
\definecolor{lightgray}{rgb}{.95,0.95,0.95}
\definecolor{lightgreen}{rgb}{0.3,0.5,0.3}
\definecolor{darkgreen}{rgb}{0.05,0.3,0.05}

\newcommand{\ra}{$\rightarrow$~}

\newtheorem{myproposition}{Proposition}
\newtheorem{myremark}{Remark}
\newtheorem{myproblemstatement}{Problem Statement}
\newtheorem{mylemma}{Lemma}
\newtheorem{mytheorem}{Theorem}
\newtheorem{mydefinition}{Definition}
\newtheorem{mycorollary}{Corollary}

\usepackage{balance}
\usepackage{hyperref}

\begin{document}

\title{Aerial Base Station Placement: \\A Tutorial Introduction}

\if\spsformat1
\name{Pham Q. Viet and Daniel Romero\thanks{Thanks to XYZ agency for funding.}}
\address{Author Affiliation(s)}
\else
\author{Pham Q. Viet and Daniel Romero\thanks{The authors are with the
    Dept. of Information and Communication Technology of the
    University of Agder, 4879, Grimstad, Norway. Email:
    \{viet.q.pham,daniel.romero\}@uia.no.\\
    This work has been funded by grant 311994 of the Research
    Council of Norway.

    © 2022 IEEE. Personal use of this material is permitted.
Permission from IEEE must be obtained for all other uses, in any current or future media, including eprinting/republishing this material for advertising or promotional purposes, creating new collective works, for resale or redistribution to servers or lists, or reuse of any copyrighted component of this work in other works.
    }}
\fi

\maketitle

\newcommand{\pomit}[1]{} 
\newcommand{\comit}[1]{\textcolor{black}{#1}} 

\begin{abstract}
\begin{bullets}%
  \blt[motivation]
  \begin{bullets}%
    \blt[purposes of ABSs]The deployment of \emph{Aerial Base
    Stations} (ABSs) mounted on board \emph{Unmanned Aerial
    Vehicles} (UAVs) is emerging as a promising technology to
    provide connectivity in areas where terrestrial infrastructure
    is insufficient or absent. This may occur for example in remote
    areas, large events, emergency situations, or areas affected by
    a natural disaster such as a wildfire or a tsunami.
    \blt[placement prob.] To successfully materialize this goal, it
    is required that ABSs are placed at locations in 3D space that
    ensure a high quality of service (QoS) to the ground terminals.
    \blt[challenges]\pomit{This problem, termed \emph{ABS placement}, is
    intrinsically non-convex and impaired by the fact that the channel
    is unknown. Besides, any potential solution needs to be adaptive
    to changes in dynamic environments. }
  \end{bullets}%
  \blt[purpose of the paper]This paper provides a tutorial
  introduction to this ABS placement problem %
  \blt[structure]%
  \begin{bullets}
    \blt[fundamental challenges]where the fundamental challenges and
    trade-offs are first investigated by means of a toy application
    example\pomit{ that, in addition, helps the reader build intuition}.
    \blt[approaches in 2D, 3D] Next, the different approaches in the
    literature to address the aforementioned challenges in both 2D
    or 3D space will be introduced
    \blt[adaptive placement] and a discussion on adaptive placement
    will be provided.
    \blt[conclusions]The paper is concluded by discussing future
    research directions.
  \end{bullets}
\end{bullets}
\end{abstract}


\section{Introduction}
\begin{bullets}%
  \blt[Motivation ABSs]
  \begin{bullets}%
    \blt[UAVs]The rapid development of the technology of
    \emph{unmanned aerial vehicles} (UAVs) has spawned a myriad of
    use cases in wireless communications.
    \blt[ABSs]One of the most prominent application scenarios involves mounting
    base stations on board UAVs to provide connectivity in areas where it is insufficient or absent.
    \blt[application examples]
    \begin{bullets}%
      \blt[remote] For example, such \emph{Aerial Base Stations}
      (ABSs) can be deployed to provide connectivity to ground
      terminals (GTs) in remote areas with no cellular
      infrastructure
      \blt[insufficient]or in mass events, such as concerts or
      festivals, where the terrestrial infrastructure is overwhelmed by
      an unusually large traffic demand.  
      \blt[damaged]Another use case involves restoring coverage after
      the ground base stations are damaged because of e.g. a terrorist
      attack or a natural disaster, such as a flood or a large fire.
      \blt[emergencies] Thanks to their high mobility and swift
      deployment capabilities, ABSs are especially useful in these
      and, in general, other emergency situations, such as traffic
      accidents or wildfires. 
    \end{bullets}
  \end{bullets}
  
  \blt[placement]
  \begin{bullets}%
    \blt[formulation] The operation of ABSs is
      illustrated in Fig. \ref{fig:network}.  The fundamental
    research problem that arises in this context is to determine where
    a collection of ABSs need to be placed in order to serve the GTs
    effectively. This problem is referred to as \emph{ABS placement}
    and has been receiving exponentially growing attention in the last
    two years. There are three main challenges that complicate this
    problem.
    \blt[challenges]
    \begin{bullets}%
      \blt[channel] First, the suitability of a location to deploy an
      ABS depends mainly on the quality of the channel to the targeted
      GTs, but this channel is unknown.
      \blt[non-convex, NP hard]
 {Second, due to the nature of the propagation of radio waves, the locations between two favorable positions need not be suitable.} In other words, typical
      performance metrics are non-convex functions of the positions of
      the ABSs, which results in non-convex  placement problems\pomit{, thereby
      causing difficulties in finding locations efficiently}.
      \blt[changes] Finally, placement algorithms need to adapt to
      dynamic environments characterized by moving GTs and
      changes in operational conditions such as the number of
      available ABSs.
    \end{bullets}
  \end{bullets}

  \begin{figure}[t!]
    \centering
    \includegraphics[width=\linewidth]{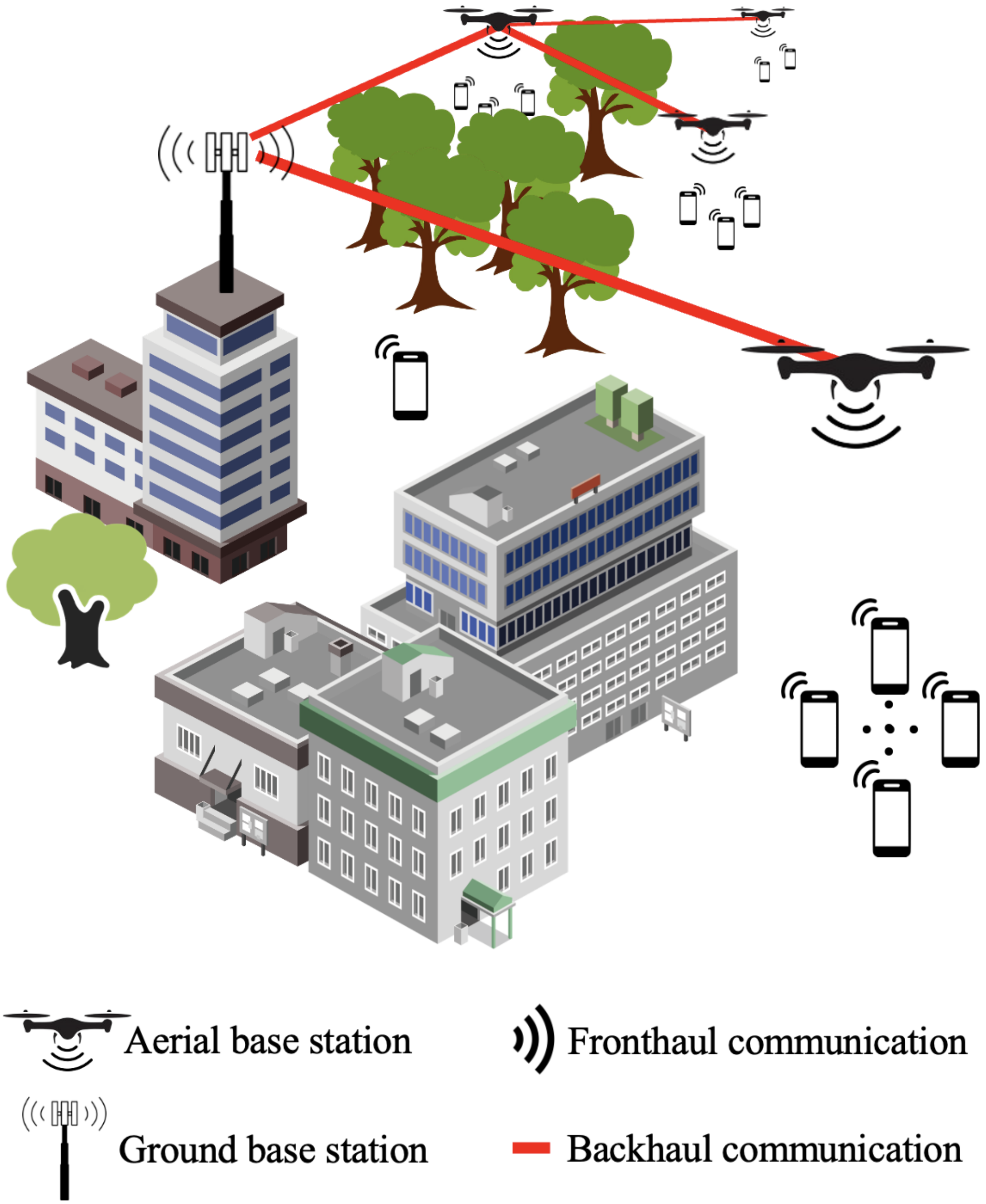}
    \caption{{ABSs connect on the one side with the GTs
        and on the other with the terrestrial infrastructure by means
        of a \emph{backhaul} connection, which can be direct or
        through other UAVs.}}
    \label{fig:network}
  \end{figure}    
  \blt[overview of the paper] 
  \begin{bullets}%
    \blt[intro] This article provides a tutorial introduction to ABS
    placement.
    \blt[rest]After building up intuition through a toy application
    example, the main existing approaches will be described
    {starting from the simplest ones and gradually
      introducing the state of the art. A taxonomy
      of methods is developed by classifying existing schemes into
      those for placement at a fixed altitude, placement in 3D space,
      and adaptive placement algorithms.}  Finally, a brief discussion
    of future directions is presented.  \pomit{
    \blt[challenges in placement] The starting point is to analyze the
    fundamental challenges and trade-offs involved in this problem by
    means of a toy example.
    \blt[2D + 3D approaches]Building upon the conclusions of this
    study, the main approaches in the literature are presented under
    different assumptions on the channel. 
    \blt[intro to adaptive] Finally, an introduction to adaptive
    placement will be provided.
    \blt[conclusion]The article concludes with some closing remarks
    and future directions.
    }
  \end{bullets}
\end{bullets}

\newcommand{\response}[1]{\textcolor{black}{#1}}

\section{Understanding the Problem}
\label{sec:problem}
\begin{bullets}%
  \blt[intro]This section investigates the fundamental challenges
  and trade-offs involved in the problem of ABS placement. To this
  end, the main elements of this problem are condensed into a simple
  toy example of ABS placement in a single dimension.

  \blt[experiment descr]
  \begin{bullets}%
    \blt[application \ra emergency on a road]Suppose that, due to
    a traffic accident or a bridge collapse, 
    \blt[GTs on a line] 10 vehicles lie static on a remote straight
    road segment of 1000 m without cellular communication
    connectivity.
    \blt[single ABS] An ABS is sent to that region to provide
    connectivity to these GTs. 
    \blt[goal]The problem is to determine the best location for the
    ABS in terms of a given \emph{quality of service} (QoS) metric,
    such as the sum rate. \response{For simplicity, it will be
      throughout assumed that a backhaul connection is available
      everywhere.
    }
    
  \end{bullets}
  \blt[Fig 1 \ra observations]
  %
  \begin{figure}[t!]
    \centering
    \begin{subfigure}[b]{\linewidth}
      \centering
      \includegraphics[width=\textwidth]{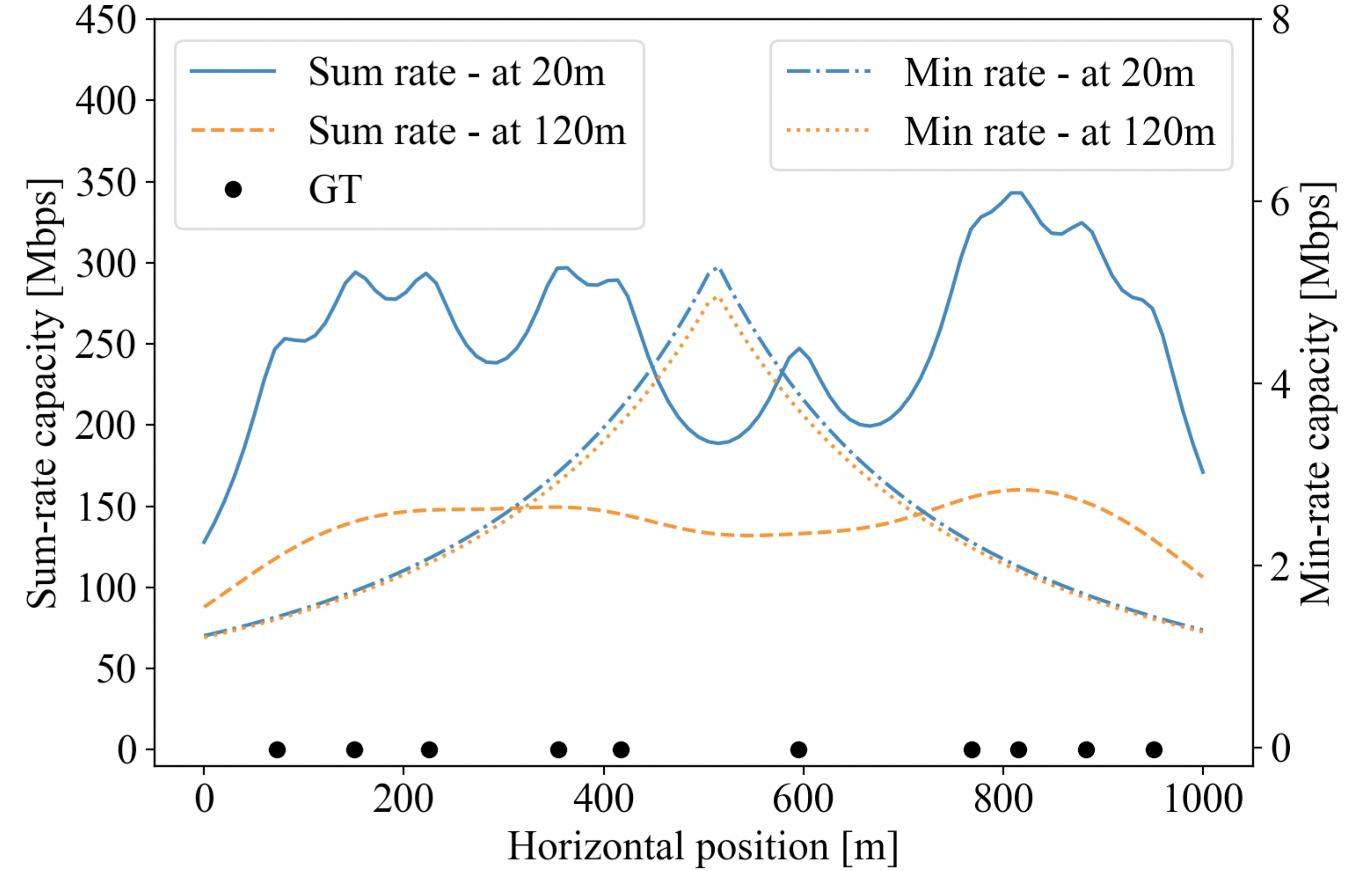}
      \caption{}
      \label{fig:positions}
    \end{subfigure}
    \begin{subfigure}[b]{\linewidth}
      \centering
      \includegraphics[width=\textwidth]{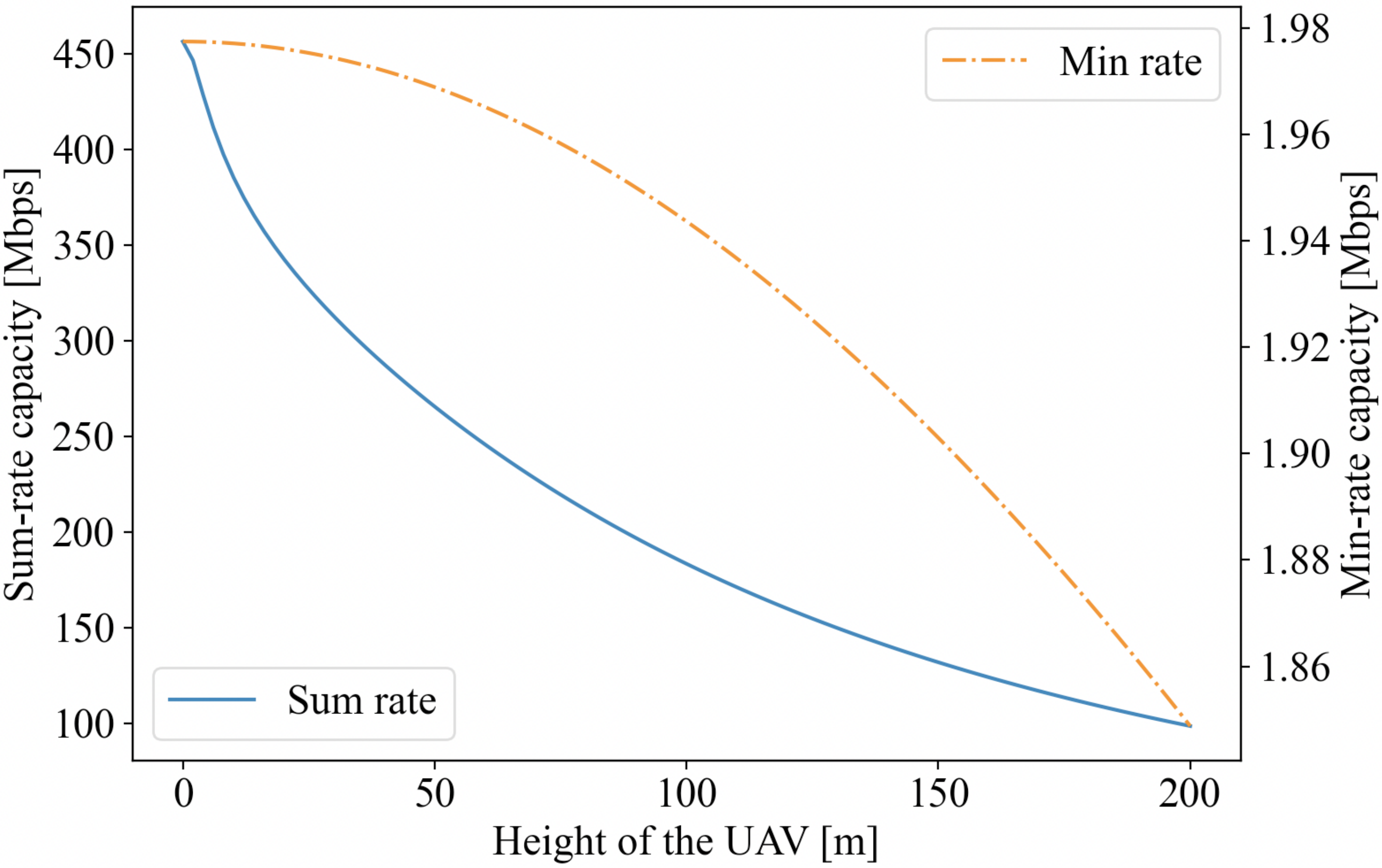}
      \caption{}
      \label{fig:heights}
    \end{subfigure}
    \caption{\color{black}  QoS metrics in an example of ABS placement of a single ABS in a single dimension: (a) sum and min rate vs. ABS position for different heights; (b) sum and min rate vs. height. 
      The horizontal
      position of the ABS in (b) is the one that maximizes the sum
      rate in (a)
      when the height is
      20~m.}
    \label{fig:qos-metrics}
  \end{figure}
  
  %
        
  \begin{bullets}%
    \blt[sum rate]Fig.~\ref{fig:positions} depicts the sum rate
    vs.~the horizontal position of the ABS placed above the road for
    different heights
    \begin{bullets}
      \blt[assumptions]%
      \begin{bullets}
        \blt[AWGN channel] assuming additive white Gaussian noise
        (AWGN) channels between the ABS and GTs
        \blt[free space] with free space propagation
        \blt[antennas] and isotropic antennas.
      \end{bullets}%
      \blt[multiple local optima]The first observation is that
      the sum rate is a rather irregular function of the
      horizontal position of the ABS with multiple local
      maxima. 
      \begin{bullets}%
        \blt[why]\pomit{To see why this is the case, note that although the
        capacity of an AWGN channel between two points in free space
        is a convex function of the distance, it can be seen that it
        is neither a convex nor concave function of the horizontal
        position of the ABS. The sum rate is the result of adding such
        non-concave functions and therefore it is natural to expect
        that it is non-concave.}
        \blt[at a low altitude]This effect is especially
        manifest for low heights (e.g. 20 m) since the
        distances between the ABS and each GT are markedly
        different. 
        \blt[at a high altitude] However, when the ABS is placed
        at a sufficiently large height  (e.g. at 120 m), these
        distances resemble each other to a greater extent and
        therefore the sum rate function becomes flatter. 
      \end{bullets}
      
      \blt[clusters]The second observation is that the sum rate tends
      to be larger when the ABS is placed on top of a \emph{cluster} of GTs
      that lie near each other, as we can observe on the right side of
      Fig.~\ref{fig:positions}. As discussed in the next section, when
      multiple ABSs need to be deployed, this idea suggests approaches
      where the GTs are grouped into clusters and each ABS is placed
      above the centroid of each cluster.
    \end{bullets}
          
    \blt[min rate]
    \begin{bullets}%
      \blt[motivation]Fig.~\ref{fig:positions} also shows the min rate
      as a function of the horizontal position of the ABS for multiple
      heights. These curves reveal that, although e.g. for a height of
      20 m the position that yields the largest sum rate would result
      in an average rate of around 350 Mbps/10 GTs $=$ 35 Mbps per GT,
      some GTs only receive around 2 Mbps. This suggests that the sum
      rate is not a satisfactory QoS metric when it comes to promoting
      fairness across GTs, as it may be necessary in certain
      applications where a minimum rate needs to be guaranteed. In
      these cases, it may be preferable to adopt the min rate as QoS
      metric.
      \blt[quasi-concavity]As per Fig.~\ref{fig:positions}, the min
      rate is a much better behaved function than the sum rate and
      features a single local maximum. Furthermore, the min rate is
      seen to be \emph{quasi-concave}, \pomit{meaning that the superlevel
      sets are convex,} which simplifies considerably its maximization.
      \blt[structure]Besides, it can be shown that the min rate to the
      left (respectively right) of its maximum equals the rate
      of the right-most (left-most) GT.
      \blt[maximum]This means that the maximum is attained where both
      of these rates coincide. In this case, the optimal horizontal
      position of the ABS is the middle point between the left-most
      and right-most GTs. More generally, in these free-space
      propagation conditions, the min rate is maximized
      when the  distance between the ABS and the farthest GT is
      minimized.
    \end{bullets}
  \end{bullets}

  \blt[Fig.~\ref{fig:heights}]
  \begin{bullets}%
    %
    \blt[decreasing]To investigate how to set the ABS altitude,
    Fig.~\ref{fig:heights} depicts the aforementioned QoS metrics
    vs. the height, where the horizontal position of the ABS is the
    one that maximizes the sum rate in Fig.~\ref{fig:positions} for
    a height of 20 m. It is observed that the sum and min rates are
    decreasing functions of the height and, therefore, maximized when
    the height of the ABS equals $0$, i.e. the ABS is on the ground.
    \begin{bullets}%
      \blt[explanation\ra Pythagoras]To understand why this is the
      case, one can resort to the Pythagorean theorem, which states
      that the distance between the ABS and a GT equals the square
      root of the square of the height of the ABS plus the square of
      the difference between the horizontal positions of the ABS and
      GT. The distance is therefore an increasing function of the
      height. Thus, when the height increases, the distance increases,
      the channel gain decreases and, therefore, the capacity
      decreases.
      \blt[free space]This seemingly counter-intuitive fact is due
      to the assumption of free-space propagation. 
      \blt[los probability] In practice, a low ABS altitude means that
      the link between the ABS and a GT will likely be obstructed by
      an object such as a vehicle or a tree. As the height
      increases, the probability of \emph{line of sight} (LoS)
      increases.
      \blt[trade-off] This gives rise to a trade-off in setting the
      altitude of the ABSs: increasing the altitude increases the
      probability of LoS but also the distance between the ABS and
      GTs~\cite{alhourani2014urban}. This implies that there exists an
      optimal altitude, as further discussed later.
    \end{bullets}
  \end{bullets}

  \blt[outlook] After having investigated the fundamental
  phenomena occurring in ABS placement by means of this toy
  example, the rest of this article will delve into approaches for
  placing multiple ABSs in different kinds of channels.
\end{bullets}

\section{Placement at a Fixed Altitude}
\label{sec:2D}

\begin{bullets}%
  \blt[free space assumption]As indicated in the previous section,
  assuming that propagation takes place in free space regardless of
  the ABS altitude implies that the optimal altitude is 0. There are
  two main ways of dealing with such an artifact:
  \begin{bullets}%
    \blt[channel model]One is not to assume free space
    propagation, as explored in the next section. 
    \blt[minimum height]The other is to fix the altitude to a
    sufficiently large value in such a way that the free-space
    assumption approximately holds.  \pomit{Setting a minimum altitude is
    further motivated by safety considerations.}

  \end{bullets}
  %
  %
  \blt[2d schemes] In this context, a large number of schemes have
  been proposed to set the 2D position of multiple ABSs in a
  horizontal plane with a given height.
  %
  \subsection{Clustering-based Placement}
  \begin{bullets}%
    \blt[K-means clustering]
    \begin{bullets}%
      \blt[Motivation]
      \begin{bullets}%
        \blt[why clustering]As discussed in the previous section, the
        sum rate tends to be high when the ABS is above a cluster of
        closely-located GTs. This suggests approaches for multiple-ABS
        placement where the GTs are grouped into clusters and an ABS is
        deployed above each one.  Each ABS therefore serves the GTs in
        its cluster.
        
        \blt[why minimize sum of distances]Given that the channel
        capacity is a decreasing function of the distance, it makes
        sense that the clustering procedure is performed in such a way
        that the GTs lie as close as possible to the assigned ABS.
        Fig.~\ref{fig:K-means} illustrates an example of placement in
        2D where this clustering task is solved using the well-known
        K-means algorithm~\cite{galkin2016deployment}. This algorithm
        groups the GTs into clusters and produces a centroid for each
        cluster. An ABS needs to be deployed above each cluster at the
        prescribed altitude. \blt[performance]{Although
          the algorithm does not provide optimal positions, its
          performance is relatively competitive with sophisticated
          placement algorithms.}

       \begin{figure}[t!]
          \centering
          \includegraphics[width=\linewidth]{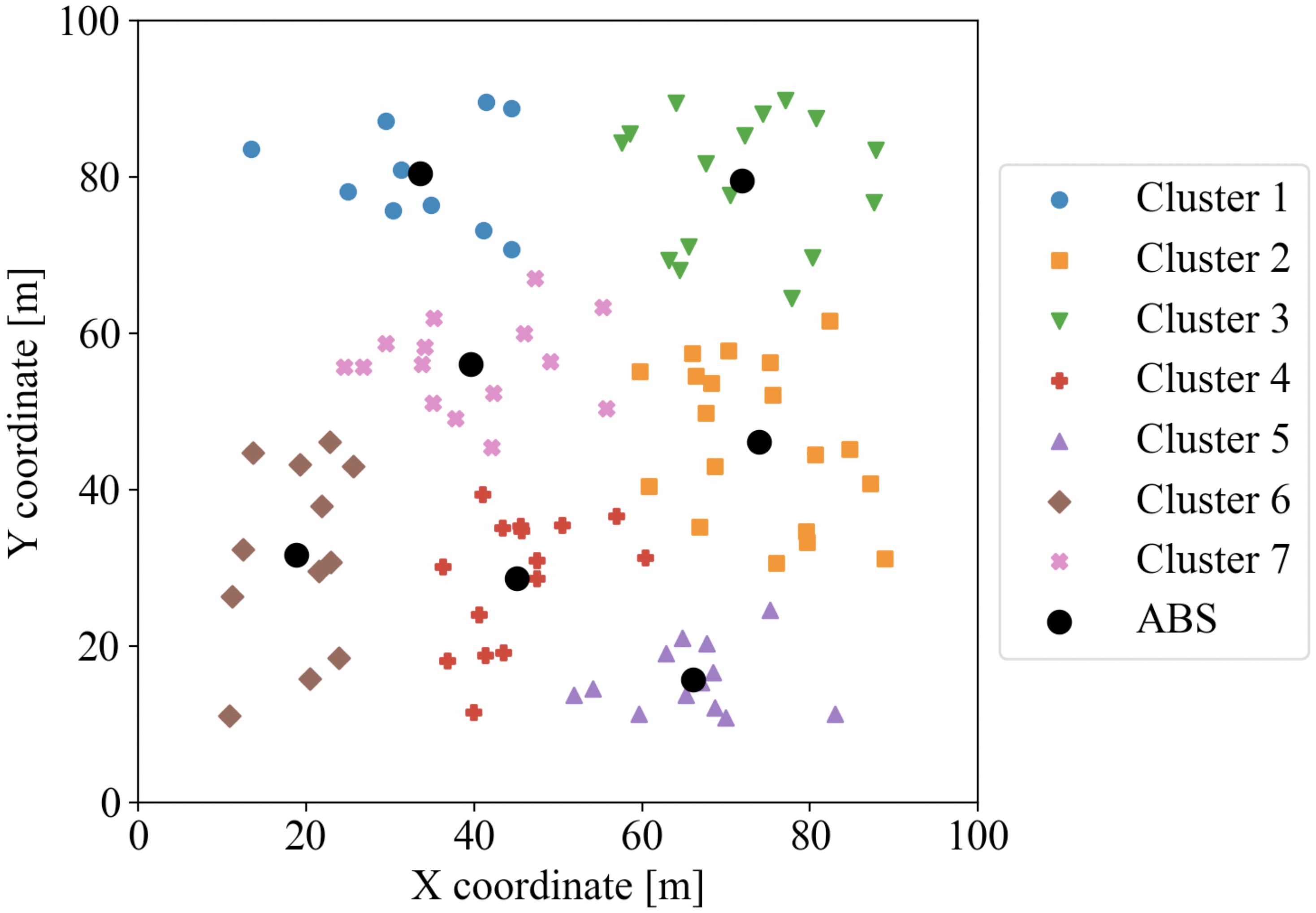}
          \caption{An application of K-means in ABS placement to group
            100 GTs into 7 clusters seen from above. Colored markers denote GTs. Black dots represents centroids where ABSs are placed.}
          \label{fig:K-means}
        \end{figure}

      \end{bullets}
      \pomit{
      \blt[problem]Finding such optimal set of locations can be shown
      to be a non-smooth, non-convex {\color{red}, and NP-hard} optimization
      problem. However, it can be solved approximately by using
      K-means with reasonable complexity.
      \blt[solution \ra k-means]
      \begin{bullets}%
        %
        \blt[descr. of K-means] Briefly, K-means
        clustering is an iterative algorithm initialized by choosing the positions of ABSs randomly. \acom{need checks on common words} The iteration process begins by
        calculating the distances between all GTs and all ABSs. GTs with a mutual nearest ABS then
        will be clustered together. Next, each cluster assigns a new position for its ABS by taking the mean of locations of its GTs. The iteration
        will be continued until there are not any new positions of ABSs appearing.  K-means is used
        not only to geographically cluster GTs but also to find \acom{favorable positions of ABSs} in each cluster. This
        minimizes the sum of distances between GTs and the ABS in each
        cluster, hence minimizing the path loss.
        \blt[figure] An application of K-means in ABS placement is
        shown in Fig.\ref{fig:K-means}. 
        %
      \end{bullets}
      }
    \end{bullets}
    
    \subsection{Circle-based Placement}
    \blt[circle-based]
    \begin{bullets}%
      \blt[motivation]
      \begin{bullets}%
        \blt[fixed num ABS]With K-means, the number of ABSs is given
        and the (sum of the squared) distances from the ABSs to the
        GTs minimized. Thus, this algorithm cannot guarantee a maximum
        distance between a GT and the nearest ABS. In practice, this
        means that a fraction of the GTs may be too far from their
        corresponding ABS to attain a certain minimum rate.
        \blt[fixed max distance]However, as indicated earlier, a
        minimum GT rate needs to be guaranteed in certain
        applications. This calls for approaches that seek the minimum
        number of ABSs required to ensure that all GTs receive at
        least a given rate. Since in free-space conditions
        guaranteeing a minimum rate amounts to ensuring a maximum
        distance, this gives rise to a formulation that is
        complementary to clustering-based placement: instead of
        minimizing the distances between GTs and their nearest ABSs
        for a given number of ABSs, the problem is now to minimize the
        number of ABSs given a maximum distance between GTs and the
        nearest ABS.
        
      \end{bullets}
      \blt[description]Since the rate decreases radially with
      distance, the set of points that receive the minimum rate from a
      given ABS forms 
      a sphere centered at the ABS. The coverage area on the
      ground is therefore the intersection between a horizontal plane
      and this sphere, which defines a circle on the ground whose
      center is the projection of the ABS on the ground.
      %
      %
      \blt[geom. problem]The ABS placement problem becomes in this way
      a geometric problem, named the \emph{geometric disc cover}
      problem, where one needs to minimize the number of circles of a
      certain radius necessary to cover a set of points, in this case
      the GT locations. 
      %
      \blt[examples in literature]Since this problem is NP-hard, a
      number of heuristics have been proposed in the literature. 
      \begin{bullets}
        \blt[lyu2017mounted]%
        \begin{bullets}
          \blt[algorithm]For example, in \cite{lyu2017mounted}, the
          ABSs are placed sequentially following an inward spiral
          around the uncovered GTs. Specifically, at each step, the
          convex hull of the locations of the uncovered GTs is
          obtained and an ABS is placed on the boundary to maximize
          the number of GTs inside its coverage circle. Afterwards,
          the newly covered GTs are marked as covered and the process
          repeated.
          
        \end{bullets}
        \blt[huang2020sparse]
        \begin{bullets}%
          \blt[descr] Another heuristic is proposed
          in~\cite{huang2020sparse}
          \blt[solution]based on a sparse-recovery optimization
          approach. The idea is to assign a \emph{virtual} ABS to each
          GT and then optimize over the locations of these virtual
          ABSs such that (i) they lie at most within the given maximum distance
          from the assigned GT and (ii) the number of different
          locations of the virtual ABSs is minimized. Once the
          optimization is completed, an \emph{actual} ABS is deployed
          for each distinct location of the virtual ABSs. Thus, by
          enforcing that many of the virtual ABS locations coincide,
          one effectively minimizes the number of actual ABSs. 
          %
          \blt[limitation \ra
          complexity] 
          The main limitation of this scheme is that the complexity of
          the optimization algorithm scales with the sixth power of
          the number of GTs and, therefore, it is only suitable for a
          small number of GTs. 
        \end{bullets}
      \end{bullets}
      
      \blt[limitations of circle-based] 
      Besides the assumption of free-space propagation, which may not
      be realistic in certain scenarios, one of the main limitations
      of circle-based approaches is that they disregard interference
      between nearby ABSs.
    \end{bullets}

    \subsection{Virtual Force-based Placement}
    \blt[virtual forces]
    \begin{bullets}%
      \blt[tradeoff]As stated earlier, the ABSs need to be
      as close as possible to their served GTs. However, in case that
      the ABSs share spectral resources, this may give rise to a large
      inter-ABS interference. For this reason, the distance between
      ABSs should be kept sufficiently high.
      \blt[virtual forces]One approach for balancing these two goals,
      namely attaining a low distance between ABSs and GTs but a large
      distance between ABSs, is by means of \emph{virtual
        forces}~\cite{andryeyev2016selforganized}.
      \blt[explanation] 
      %
      Particularly, one can consider an attractive force for each (ABS,GT)
      pair and a repulsive force for each (ABS,ABS) pair. 
      \begin{bullets}%
        \blt[attractive]Specifically, for each (ABS,GT) pair, the
        attractive force acts on the ABS in the direction towards the
        GT, whereas
        \blt[repulsive]for each (ABS,ABS) pair, the repulsive force
        acts on both ABSs in the direction away from each other. 
      \end{bullets}
      Each ABS computes the sum of the virtual forces that act on it
      and moves in its direction. In this way, the ABSs tend to spread
      across space and, at the same time, move to hotspots with a high
      number of GTs.

      %
      %
      \blt[advantage] The strengths of the approach based on virtual
      forces in \cite{andryeyev2016selforganized} are low complexity
      and adaptability to changes in the operational environment.
      %
      %
    \end{bullets}
  \end{bullets}

\end{bullets}
  
\section{3D ABS Placement}
\label{sec:3D}
Relative to the schemes in the previous section, it is clear that an
improved QoS can be attained if the ABS altitude is also optimized. As
concluded earlier from the toy example, this requires lifting the
free-space propagation assumption. Each of the ensuing subsections
outlines a different possibility towards this end.




\subsection{Channel-agnostic Placement}
\begin{bullets}%
  \blt[general principle]
  \begin{bullets}%
    \blt[descr]Instead of adopting any assumptions on the channel, one
    can think of relying on measurements. A simple possibility is that
    each GT measures the channel to each ABS and associates with the
    one whose channel is strongest. In this way, a large number of GTs
    will be associated with those ABSs at locations with favorable
    propagation conditions. This information can be used to determine
    at which locations one needs to deploy a larger number of ABSs.
    
    \blt[strengths]
    \begin{bullets}%
      \blt[no location/detailed ch. est]The main strength of such an
      approach is that ABSs need not know the locations of or
      distances to the GTs. A detailed knowledge of the channel is not
      required either: it suffices to know which is the strongest one,
      which can be known by the GTs based on the received signal
      strength (RSS) of beacon signals transmitted by the ABSs.
      \blt[less communication overhead] Therefore, this scheme
      features a low communication overhead between ABSs and GTs.
    \end{bullets}
  \end{bullets}
  
  \blt[example work] An example of channel-agnostic placement can be found in \cite{park2018formation}.
  \begin{bullets}%
    \blt[descr]The scheme therein introduces several classes of
    virtual forces to set the 3D positions of the ABSs.
    \begin{bullets}%
      \blt[h-position]Some of these forces are repulsive and aim at
      spreading the ABSs across space. Others are attractive and bring
      together ABSs that have a different number of associated GTs.
      \blt[altitude]The altitude is set by another virtual force that
      aims at decreasing the coverage area of overloaded ABSs by
      increasing their height. The obvious drawback of such an
      approach is that the channel to the rest of GTs will generally
      worsen.
    \end{bullets}
    \blt[strengths]\pomit{\acom{omit} 
    The main strength of this work is that it can be implemented in a
    decentralized fashion.} 
    \end{bullets}
    
  \blt[limitations]
    \begin{bullets}%
      \blt[computation] The main limitation of channel-agnostic
      placement is that it cannot know whether a spatial arrangement
      of ABSs is satisfactory beforehand, i.e., before the ABSs
      actually occupy those locations, which is clearly inconvenient
      if one wishes to find a close-to-optimal placement. 
      \blt[good locations]Besides, there may be locations with very
      favorable propagation conditions that are never discovered by
      the system because no ABS ever visits them. 
      \blt[too coarse]\comit{Moreover, by relying solely on the number of
      associated GTs with each ABS, it is expected that the resulting
      placement arrangements yield lower performance than if more
      detailed information were used, such as  channel state information
       or location  information.}
    \end{bullets}
\end{bullets}

\subsection{Altitude Optimization via Empirical Models}
\begin{bullets}%
  \blt[motivation modeling]
  \begin{bullets}%
    \blt[limit. ch. agnostic]As indicated earlier, channel-agnostic
    schemes cannot determine how favorable a candidate ABS location is
    unless an ABS actually visits that location and the channels to
    the GTs are measured.
    \blt[modeling]A natural approach to alleviate this limitation is
    to rely on a  model that provides the channel quality given
    the locations of the ABSs and GTs. However, as discussed
    previously, such a model should not assume free-space propagation.
    
  \end{bullets}

  \blt[model in alhourani2014urban]The most widely used model is the
  one in~\cite{alhourani2014urban}, which captures the phenomenon
  described earlier that the probability of LoS increases with
  elevation. 
  %
  \begin{bullets}%
    \blt[LOS vs. NLOS]Specifically, this empirical model classifies
    links between a GT and an ABS into one out of two categories,
    depending on the magnitude of the excess path loss relative to
    the free-space path loss, namely LoS (or near-LoS) and non-LoS.
    \blt[PL distribution]The path loss is then modeled as a Gaussian
    distribution whose parameters depend on the elevation of the link,
    the link category, and characteristic parameters of the
    environment (e.g. urban or rural).
    \blt[setting params] The values of these characteristic parameters
    could be found by fitting a set of measurements. In
    \cite{alhourani2014urban}, reference values are provided by
    fitting simulated data in four scenarios that adhere to the
    guidelines of the International Telecommunication Union (ITU-R).
    \pomit{,
    which condense the propagation characteristics of an environment
    into three parameters, namely the ratio of the area of building to
    the total area (dimensionless), the average number of buildings on
    a unit area (buildings/$\rm{km^{2}}$), and a scale parameter
    describing the height distribution of the buildings}
    
    \blt[Prob. LOS]The model also provides the probability of LoS as
    a function of the elevation of the link. Similarly, the
    parameters of this function are also fitted to the data. 
    \blt[mean PL]Knowing the probability of each category as well as
    the distribution of the path loss under each of them allows the
    computation of the \emph{mean} path loss. 
  \end{bullets}
  
  \blt[placement approaches]
  \begin{bullets}%
    \blt[formulation\ra optimization]A large number of works rely on this
    empirical model. The typical approach involves formulating an
    optimization problem that aims at optimizing a certain QoS
    metric subject to communication constraints where the optimization
    variables are the locations of the ABSs.
    \blt[mean PL]The positions are related to the necessary
    communication metrics, such as capacity, by approximating the path
    loss between a GT and an ABS as the \emph{mean} path loss given by
    the aforementioned model.
    \blt[non-convex]The difficulty is that, given the form of the
    model, the resulting optimization programs are non-convex. 
    \blt[examples]
    \begin{bullets}%
      \blt[kalantari2016number]This has motivated global optimization
      approaches such as the one in~\cite{kalantari2016number}, which
      relies on particle swarm optimization.  Here, the problem is to
      minimize the number of ABSs to ensure a minimum average spectral
      efficiency.       
      %
      \blt[shehzad2021backhaul]
      \blt[liu2019deployment] Similarly, genetic algorithms are
      applied      in~\cite{shehzad2021backhaul,liu2019deployment}. { Remarkably, in \cite{liu2019deployment}, a K-means rule based
        on a genetic algorithm is developed to cluster GTs and ABSs
        are trained via Q-learning to find favorable 3D positions and
        serve roaming ground users.}
      \blt[hammouti2019mechanism] Another example is
      \cite{hammouti2019mechanism}, which provides an \emph{ad hoc} algorithm
      where the height is optimized \pomit{in a decentralized fashion} using a
      game-theoretic approach. \comit{The objective is to maximize the sum
      rate under different constraints such as the requirement of
      minimum spectral efficiency of each associated GT.}
    \end{bullets}%
  \end{bullets}%
\end{bullets}

\subsection{Placement Using Terrain Maps}
\begin{bullets}%
  \blt[motivation]
  \begin{bullets}%
    \blt[emp. models \ra limitations] The approaches considered in the
    previous section rely on empirical models of the \emph{mean} of
    the path loss across a class of scenarios, e.g.  generic urban
    environments. However, the actual value of the path loss may
    greatly differ from its mean, which suggests that schemes
    based on such empirical models may yield highly
    sub-optimal placements in a specific environment.
  \end{bullets}
  \blt[description] One possible approach to accommodate information
  about the path loss in a specific propagation environment is to rely
  on  
  \begin{bullets}%
    \blt[Terrain map/3D model] terrain maps or 3D city models; 
    %
    see Fig.~\ref{fig:example-placement}.
    \begin{figure}[t!]
      \centering
      \includegraphics[width=0.9\linewidth]{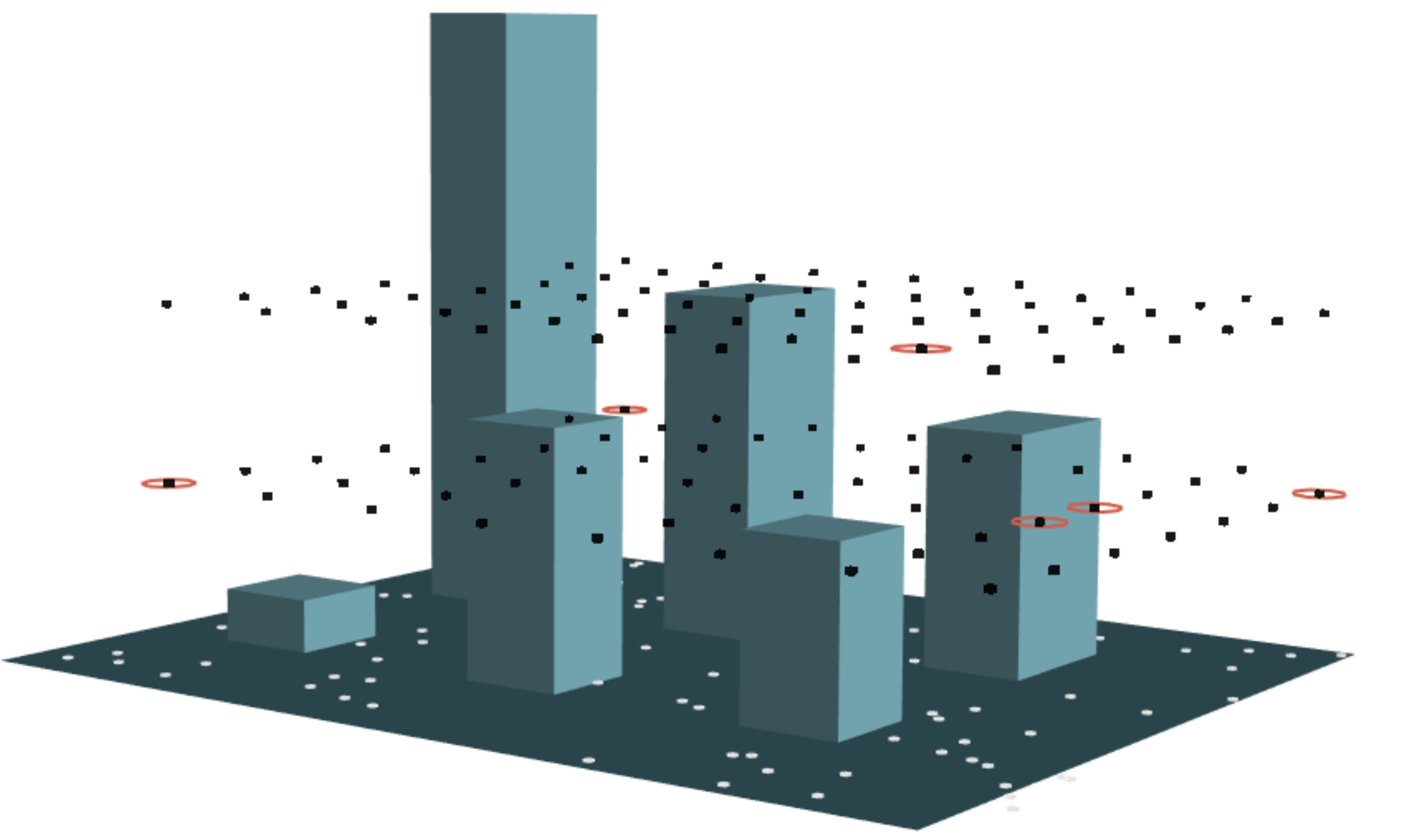}
      \caption{Example of {a 3D city map}. Markers on the
        ground denote GTs whereas orange circles stand for ABS
        positions. \response{Black dots in the air represent a fly grid and the
        placement criterion is to minimize the number of deployed ABSs
        while guaranteeing a minimum rate at each GT; see section on
        radio maps.}  }
      \label{fig:example-placement}
    \end{figure}
    \blt[how use terr. maps]Techniques such as ray tracing can be then
    used to predict the actual path loss between the GTs and positions
    that no ABS has necessarily ever visited.     
    \blt[related works]
    \begin{bullets}
      \blt[sabzehali2021orientation]A related approach is pursued
      in~\cite{sabzehali2021orientation}, where a 3D city map is used
      to construct a set of cones in such a way that if a GT is in the
      cone associated with an ABS location and vice versa, then they
      can communicate with LoS. An optimization problem is then
      formulated and the  solution approximated with a greedy
      algorithm. 
      \blt[qiu2020reinforcement]Another possibility could be to attempt a 3D extension of~\cite{qiu2020reinforcement}, where
      a 
      {relay algorithm based on a double deep Q-network with prioritized experience }
      is applied to maximize the number of covered GTs using path loss predictions based on a 3D city map.
    \end{bullets}

  \end{bullets}
  \blt[strength] Clearly, the fact that 3D models or terrain maps enable predictions
  of the channel from the GTs to each spatial location without
  visiting it with an ABS  constitutes  an important
  advantage over the channel-agnostic approach alluded to previously. 
  This is vitally important to 
  decrease the time required by placement algorithms. 
  
  \blt[limitations]
  \begin{bullets}%
    \blt[rarely available] Unfortunately, terrain maps or 3D models
    are seldom available and, even when they are, their resolution is
    typically insufficient for predicting the channel in conventional bands.
    \blt[high resolution] To see this, note that a typical resolution
    for a terrain map is~20~m~\cite{sabzehali2021orientation}. This is
    much larger than the scale of channel variations, usually in the order of the
    wavelength, which typically ranges from a few millimeters to a few
    centimeters.
    \blt[inside building] Besides, approaches based on this kind of
    maps cannot accommodate the case where a user is inside a
    building. \pomit{Fortunately, this challenge can be overcome by using
    radio maps, as discussed next.}
  \end{bullets}
\end{bullets}

\subsection{Placement Using Radio Maps}
\label{sec:radiomaps}

\begin{bullets}%
  \blt[Motivation]Terrain maps and 3D models provide the positions and
  shapes of obstacles in the propagation environment. Therefore, they
  provide the path loss in an \emph{indirect} fashion. This
  observation suggests that it can be beneficial to directly map path
  loss.
  \blt[Radio maps]
  \begin{bullets}%
    \blt[what is radio map]Specifically, one can utilize radio maps
    that provide the shadowing attenuation between all possible ABS
    locations and the GTs.
    \blt[difference + strength compared to 3D city maps] As an
    important advantage relative to city maps, radio maps provide
    channel information about the environment even when GTs lie inside
    buildings.
    
    \blt[obtaining radio maps]Radio maps are typically based on the
    so-called \emph{radio tomographic model}, which prescribes that
    the attenuation between two points equals the line integral of a
    function of the spatial location termed \emph{spatial loss field}
    (SLF), which quantifies how much a radio signal attenuates at each
    location. By gathering measurements at a collection of pairs of
    points, one can estimate the SLF and then use it to predict
    attenuation between any pair of points.
    
    \blt[discretization]To be able to manipulate this line integral,
    the SLF needs to be approximated by a piecewise constant function,
    meaning that only its values at a set of voxels are stored. To
    obtain the line integral, one needs to determine which voxels the
    line segment between the ABS and a GT traverses and take a linear
    combination of the values of the SLF at those voxels. An
    illustration in 2D is shown in Fig.~\ref{fig:radio-map}.

    
    %
    \begin{figure}[t!]
      \centering
      \includegraphics[width=\linewidth]{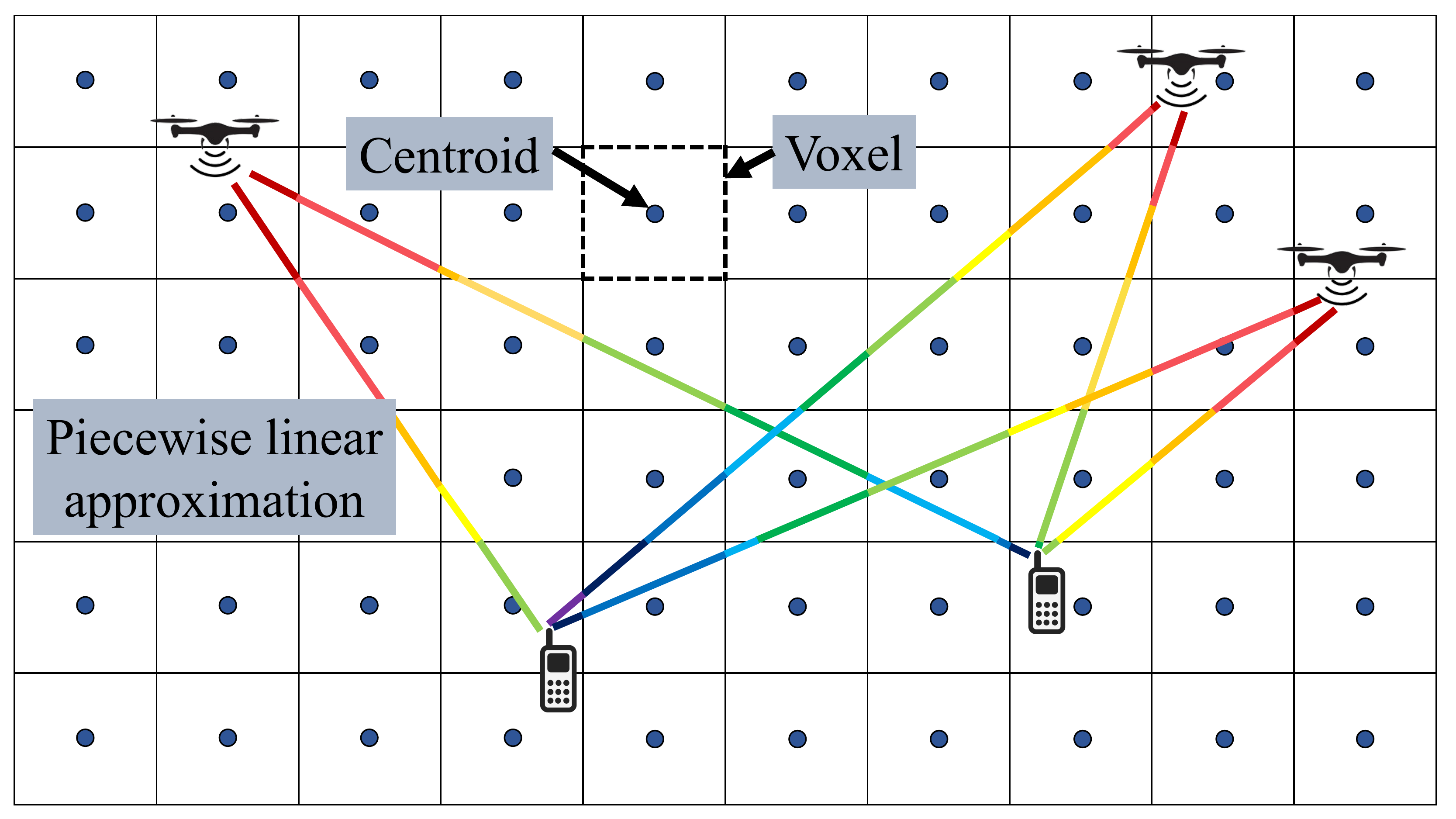}
      \caption{In the tomographic model, the attenuation between two
        points is a function of the line integral of the SLF.}
      \label{fig:radio-map}
    \end{figure}
    %
    %

  \end{bullets}
  \blt[Sparse placement]Radio maps have been used for 3D
  ABS placement in \cite{romero2022aerial}.
  \begin{bullets}%
    \blt[problem] In this work, the set of all possible ABS
    locations is discretized into a grid and the path loss from each
    GT to each grid point is obtained by means of a radio map. The
    placement problem is then formulated as finding the smallest
    number of ABSs required to serve all GTs.
    \blt[strengths]One of the main strengths is that the resulting
    optimization problem is convex and its solution entails just
    linear complexity. Besides, this scheme can take into account
    no-fly zones or airspace occupied by buildings, unlike the vast
    majority of placement algorithms in the literature.
    \blt[limitations] The limitation of this kind of schemes is that
    the construction of a high-resolution radio map could be
    challenging if a fine discretization of the 3D space is required.
  \end{bullets}

  \blt[comparison]To close this section, a representative subset of
  the algorithms discussed so far are compared in
  Fig. \ref{fig:experiment}
  . The simulation takes place in an
  urban environment such as the one in
  Fig.~\ref{fig:example-placement}. {As can be seen, the
    placement algorithm based on radio maps \cite{romero2022aerial}
    requires the smallest number of ABSs thanks to its awareness of
    the path loss in the operational region.}
  \begin{figure}[t!]
    \centering
    \includegraphics[width=\linewidth]{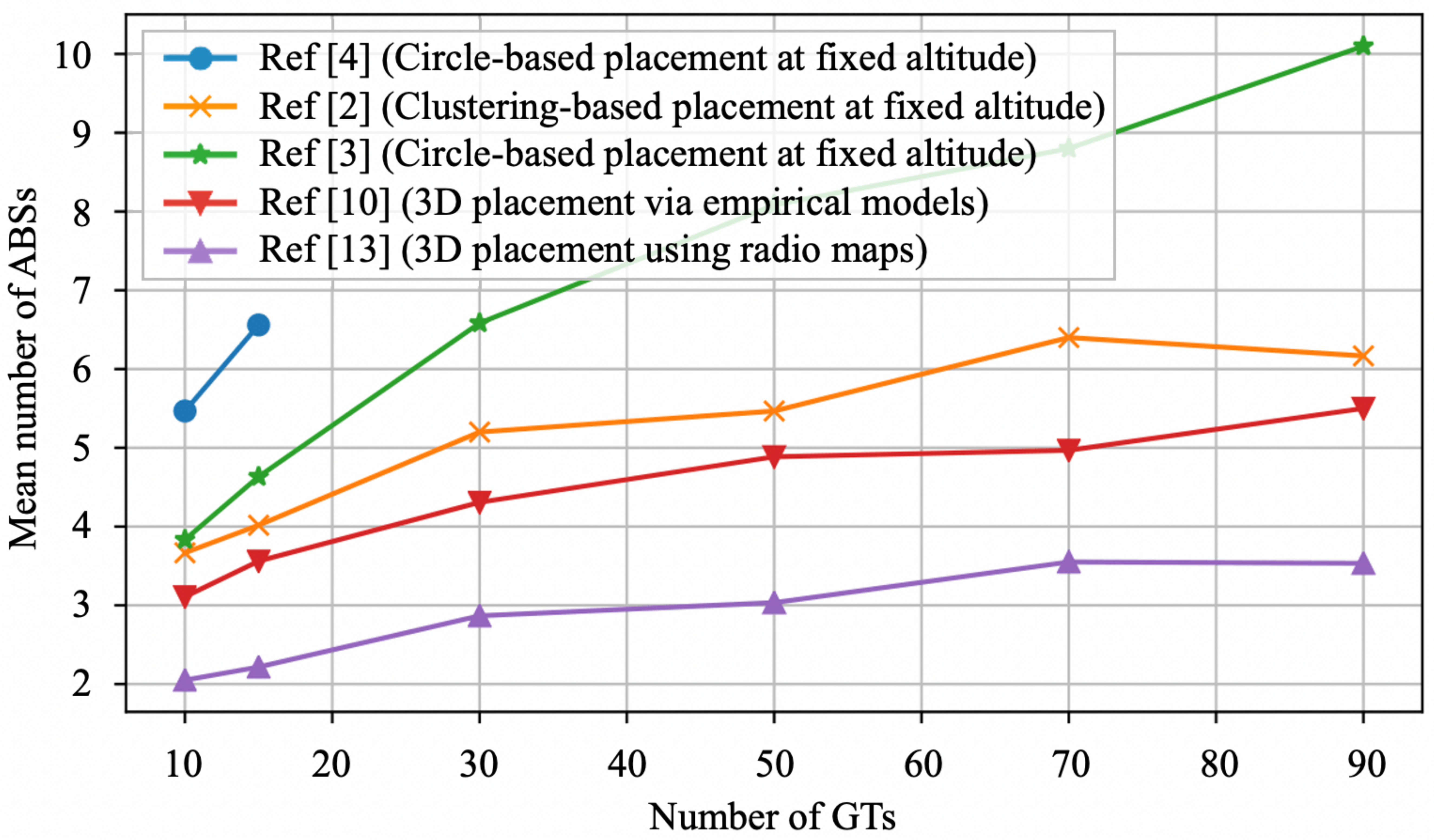}
    \caption{Comparison of a representative subset of placement algorithms. {The placement criterion is to minimize the number of required ABSs to guarantee a minimum rate at all GTs.}}
    \label{fig:experiment}
  \end{figure}
  {Therefore, using radio maps yields a clear advantage
  over schemes which adopt more simplistic assumptions about the channel.}

\end{bullets}  
\section{Adaptive Placement}
\label{sec:adaptive}
\begin{bullets}%
\blt[intro]
\begin{bullets}%
  \blt[motivation from mentioned works] The approaches introduced so
  far cannot naturally cope with changes in the positions of the ABSs
  and GTs over time.
  %
  %
  \blt[attention on adaptive]This motivates schemes where the ABSs can
  adapt their positions to changes either in the GT locations or
  in the working conditions. For example, if an ABS becomes inoperative,
  e.g. because it runs out of battery, the remaining ABSs need to 
   adapt their positions accordingly.
   \blt[overview] \pomit{This section introduces adaptive placement.}
\end{bullets}

\blt[example works] 
\begin{bullets}%
  \blt[sector-based] 
  \begin{bullets}%
    \blt[description]One simple approach inspired
    by~\cite{wang2018adaptive} could rely on simple ad-hoc
    adaptive algorithms. For the sake of illustration, consider a 1D
    setup with a single ABS that has a right and a left sector, each
    one with an antenna that points in either direction. Each
    GT connects to the sector from which it receives the highest
    power.  The ABS can then move right if more GTs are connected to
    the right sector and vice versa. This idea could also be extended
    to placement in 2D by using at least 3 sectors.
    \blt[strengths]One of the main strengths of such an approach would
    be that no knowledge of the channel or GT positions would be necessary. 
    \blt[limitations]However, applying such an idea to multiple ABSs
    and studying the convergence properties remains an open problem.

  \end{bullets}
  \blt[stochastic optimization]These difficulties can be sidestepped
  by resorting to the framework of stochastic
  optimization. Specifically, adaptive placement for multiple ABSs in
  2D space is studied in \cite{romero2019noncooperative}. %
  \begin{bullets}
    \blt[utility function]There, multiple ABSs adapt their locations
    by moving in the direction of a gradient estimate of a suitably
    designed utility function. The movement of the users leads to a
    change in this function but these changes can be naturally tracked
    by the stochastic optimization algorithm. 
    \blt[strengths]The main strength of such an approach is that ABSs
    need not communicate with each other or with any central
    controller: the scheme is decentralized and non-cooperative.
    %
    %
    \blt[limitations]The downside is that the location of the GTs and
    a channel model are required. 
  \end{bullets}
  
  \blt[reinforcement] Finally, more sophisticated approaches could be
  developed based on reinforcement learning, e.g. along the lines of
  \cite{liu2019deployment}. The main challenge with such approaches
  is to avoid the need for retraining every time the operational
  conditions, such as the environment or number of ABSs, change.  

\end{bullets}
\end{bullets}

\section{Conclusions and Open Problems}
\label{sec:conclusion}

\begin{bullets}
  \blt[conclusions]%
  \begin{bullets}
    \blt[intro]This article provided a tutorial introduction to ABS
    placement in UAV-assisted networks.
    \blt[understanding the problem] First, a toy example shed light on
    the fundamental phenomena occurring in this problem. It was observed
    that common QoS metrics may have multiple local optima and that the
    optimal altitude under the free-space propagation assumption is 0.
    \blt[2D, 3D] Next, different approaches for 2D and 3D placement were
    presented according to how they deal with the channel and 
    \blt[Adaptive]adaptive implementations briefly discussed. 
  \end{bullets}

  \blt[Open problems]Although ABS placement has been subject to
  extensive research efforts, a number of open problems still remain.
  \begin{bullets}
    \blt[Adaptivity]First, most algorithms are non-adaptive. This implies that small changes in the operational
    conditions or GT locations may result in large changes in the ABS locations or even the number of ABSs. This calls for adaptive alternatives which can furthermore account for the physical constraints on the movement
    of the UAVs.
    {\blt[Battery \ra operational time] Likewise, little
      attention has been paid to the fact that the operational time of
      ABSs is finite due to the limited capacity of their batteries or
      energy sources.
    \blt[Fixed wing]To alleviate this limitation, schemes for fixed-wing UAVs would be desired.
    }%
    {
    \blt[multi-objective optimization]%
    \begin{bullets}%
      \blt[motivation]Besides, it would also be convenient to develop
      schemes for      
      joint optimization of key performance indicators such as 
      \blt[example]throughput, time delay, and spectral efficiency. 
      %
      %
      \blt[solution]
    \end{bullets}%
    }%
    
    {\color{black} \blt[a comprehensive tutorial] Last but not least,
      as the number of works in this active research area grows, a
      comprehensive survey that complements the present introductory
      article will become highly convenient.  }
    
  \end{bullets}
\end{bullets}

~\\[.3cm]
\noindent\textbf{Pham Q. Viet} received his B.Sc. and M.Sc. degrees in Telecommunications Engineering from Ho Chi Minh City University of Technology, Viet Nam, in 2018 and 2020, respectively. He is currently a Ph.D. student with the Department of Information and Communication Technology, University of Agder, Norway. His research interests mainly lie in the areas of machine learning, optimization, signal processing, and aerial communications.

~\\[.3cm]
\noindent\textbf{Daniel Romero} (M’16) received his M.Sc. and Ph.D. degrees in Signal Theory and Communications from the University of Vigo, Spain, in 2011 and 2015, respectively. He is currently an associate professor with the Department of Information and Communication Technology, University of Agder, Norway. He also serves as an editor for Elsevier Signal Processing. Previously, he was a post-doctoral researcher with the Digital Technology Center and Department of Electrical and Computer Engineering, University of Minnesota, USA. His main research interests lie in the areas of machine learning, artificial intelligence, optimization, signal processing, and aerial communications.


\balance
\printmybibliography
\end{document}